\documentclass{article}
\usepackage{pifont}

\usepackage[preprint]{corl_2026} % Use this for the initial submission.

\usepackage{cite}
\usepackage{amsmath,amssymb,mathrsfs,amsfonts,bm}
\usepackage[ruled,vlined]{algorithm2e}
\usepackage{graphicx}
\usepackage[caption=false, font=footnotesize]{subfig}
\usepackage{multirow}
\usepackage{siunitx}
\usepackage{xcolor}
\usepackage{booktabs}
\usepackage{wrapfig}

\newcommand{\ad}{\mathrm{ad}}

\title{Do Rigid-Body Simulators Dream of Soft Robots? Learning Contact-Rich Manipulation for Tendon-Driven Continuum Robots}

\author{
  Chengnan Shentu, Nicholas Baldassini, Tongjia Zheng, Priyanka Rao, Jessica Burgner-Kahrs\\
  University of Toronto, Canada\\
  \texttt{Project Page:} \url{https://continuumroboticslab.github.io/opencr-mujoco/} \\
}

\begin{document}
\maketitle

%===============================================================================
% FIGURE 1
%===============================================================================
\begin{figure}[h]
  \centering
  \includegraphics[width=13.5cm]{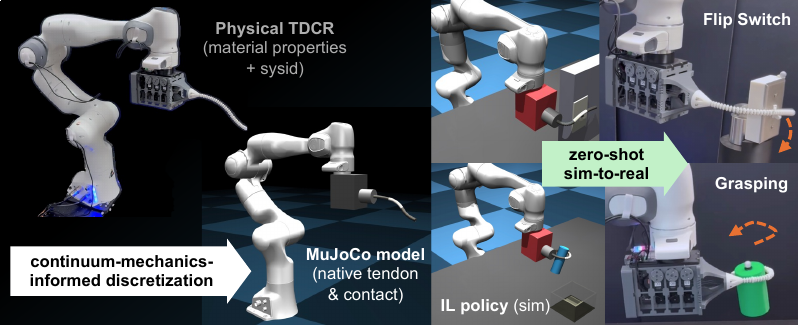}
  \caption{\textbf{Sim-to-real pipeline for contact-rich manipulation with tendon-driven continuum robots.} A continuum-mechanics-informed discretization places the TDCR natively inside MuJoCo, unifying tendon forces, body contact, and dynamics in a single physics pipeline. Policies are trained from teleoperated demonstrations in simulation and deployed zero-shot to the real robot.}
  \label{fig:fig1_overview}
\end{figure}

%===============================================================================
% ABSTRACT
%===============================================================================
\begin{abstract}
Learning contact-rich, whole-body manipulation for soft continuum robots is held back by the lack of simulation infrastructure that has accelerated rigid-robot manipulation. Existing soft robot simulators are physically grounded but lack the contact handling, actuation support, or learning integration needed for contact-rich manipulation; rigid-body approximations offer these capabilities but sacrifice physical grounding. We bridge this gap for tendon-driven continuum robots (TDCRs) by deriving a continuum-mechanics-informed discretization that places the soft robot natively inside MuJoCo, unifying tendon forces, body contact, and dynamics in a single physics pipeline. We validate the simulator against a Cosserat rod reference (static and dynamic) and real TDCR hardware. We then train state-based imitation learning policies via teleoperation in simulation and deploy them zero-shot to a physical 3-segment TDCR on a 7-DoF Franka arm across two contact-rich manipulation tasks. To our knowledge, this is the first demonstration of sim-to-real transfer for contact-rich manipulation with continuum robots.
\end{abstract}

\keywords{Simulation, Soft Robotics, Sim-to-Real Transfer.}

%===============================================================================
% 1. INTRODUCTION
%===============================================================================
\section{Introduction}
\label{sec:introduction}

Contact-rich manipulation has seen rapid progress driven in part by mature simulation infrastructure~\citep{todorov_tassa_2012_mujoco, akkaya2019solving, tang2023industreal}. Compliance is ubiquitous in this success: from rubber-padded fingertips, to compliant parallel jaw grippers, to deformable tactile sensors, compliance at the point of contact is consistently critical for robust manipulation. Soft continuum robots take this principle further, distributing active compliance along the entire body. Despite this, progress toward whole-body soft manipulation remains limited. 
Soft continuum robots face several challenges for learning in the real world: their high-dimensional configuration spaces demand careful calibration and extensive demonstration data, iterating on physical hardware is slow and resource-intensive, and their full shape is difficult to observe. Simulation can address all three, yet no simulator combines the capabilities needed for contact-rich manipulation learning: physical grounding in continuum mechanics, a contact solver, faithful actuation modeling, computational efficiency, and learning pipeline integration (Table~\ref{tab:simulator_comparison}).

Existing continuum-mechanics simulators~\citep{mathew_renda_2022_sorosim, gazzola_mahadevan_2018_forward_inverse, choi_jawed_2024_dismech} are physically grounded and some achieve robust contact resolution~\citep{choi_jawed_2024_dismech}, but they are specialized frameworks designed for modeling and analysis: actuation is modeled abstractly (e.g., as distributed loads or natural curvature changes) rather than at the level of physical inputs like tendons, leaving the user to derive a command mapping and omitting the inter-tendon coupling and tension-induced stiffness that physical tendons contribute to system dynamics. There is also no support for complex multi-object scenes or learning pipelines. Coupling a continuum solver with an existing rigid-body engine~\citep{kasaei_mohsen_2024_softmanisim} addresses some of these gaps, but current implementations only handle known external loads on the soft body, not contacts that emerge from interaction with the environment. Pseudo-rigid-body models~\citep{graule_wood_2022_somogym} sidestep these issues by placing the soft robot directly inside a physics engine, gaining native contact and interaction with rigid robots and objects. However, model parameters are empirically calibrated rather than derived from physics, with no convergence guarantee and no assurance that the resulting simulation generalizes beyond the calibration conditions. We show that the pseudo-rigid-body approach can be made physically grounded: joint stiffnesses are derived analytically from the rod's material properties, and the discretization provably converges to the continuous rod at second order. Our contributions are:

\textbf{Simulation:} We derive a continuum-mechanics-informed discretization that places tendon-driven continuum robots (TDCRs) natively inside MuJoCo with no engine modifications, achieving physical grounding, contact resolution, native tendon actuation, computational efficiency, and learning pipeline integration. We validate against a Cosserat rod reference and real hardware (Sec.~\ref{sec:sim_validation}--\ref{sec:real_validation}).

\textbf{System:} We show this simulation approach unlocks the standard sim-to-real imitation learning workflow (teleoperated data collection, imitation learning, and zero-shot policy deployment) on a 9-tendon, 3-segment TDCR mounted on a 7-DoF Franka arm. To our knowledge, this is the first sim-to-real transfer of contact-rich manipulation policies for soft continuum robots.

%===============================================================================
% 2. RELATED WORK
%===============================================================================
\section{Related Work}
\label{sec:related_work}

%===============================================================================
% TABLE
%===============================================================================
\begin{table}[h]
\centering
\vspace{-3mm}
\caption{Comparison of soft continuum robot simulators across the five capabilities required for learning contact-rich manipulation.}
\label{tab:simulator_comparison}
\small
\setlength{\tabcolsep}{2.5pt}
\begin{tabular}{@{}lcccccccc@{}}
\toprule
 & \scriptsize SOFA & \scriptsize Shooting & \scriptsize SoRoSim
 & \scriptsize Elastica & \scriptsize DisMech & \scriptsize SoftManiSim
 & \scriptsize SoMo & \scriptsize \textbf{Ours} \\
 &\scriptsize\citep{faure_cotin_2012_sofa} &\scriptsize\citep{till_rucker_2019_realtime_dynamics} & \scriptsize\citep{mathew_renda_2022_sorosim} &  \scriptsize\citep{naughton_gazzola_2021_elastica} & \scriptsize\citep{choi_jawed_2024_dismech} & \scriptsize\citep{kasaei_mohsen_2024_softmanisim} &  \scriptsize\citep{graule_wood_2021_somo} & \\
\midrule
Modeling basis
 & \scriptsize FEM & \scriptsize Cosserat & \scriptsize Cosserat 
 & \scriptsize Cosserat & \scriptsize Kirchhoff & \scriptsize Cosserat 
 & \scriptsize $\triangle$Empirical & \scriptsize Kirchhoff \\
Contact solver         & \checkmark & $\times$   & $\times$   & $\triangle$\textsuperscript{a} & \checkmark & $\times$\textsuperscript{b}   & \checkmark & \checkmark \\
Actuation modeling     & \checkmark & $\triangle$\textsuperscript{c} & $\triangle$\textsuperscript{c} & $\checkmark$   & $\times$\textsuperscript{d}   & $\triangle$\textsuperscript{c} & $\times$\textsuperscript{e}   & \checkmark \\
Computational efficiency      & $\times$  & \checkmark & \checkmark & $\triangle$\textsuperscript{f} & \checkmark & \checkmark & \checkmark & \checkmark \\
Learning integration   & $\triangle$\textsuperscript{g} & $\times$   & $\times$   & \checkmark   & $\times$   & \checkmark & \checkmark & \checkmark \\
\bottomrule
\end{tabular}

\vspace{2pt}
{\raggedright\small
\textsuperscript{a}Penalty-based; explicit integration limits stability for stiff rods.
\textsuperscript{b}Contact must be specified as known external loads, not resolved by a contact solver.
\textsuperscript{c}Tendons modeled as user-supplied distributed loads.
\textsuperscript{d}Prescribed as changes to the rod's rest curvature, no physical actuator is modeled.
\textsuperscript{e}Fixed moment at each joint, independent of configuration.
\textsuperscript{f}Explicit time integration limits real-time performance for stiff materials.
\textsuperscript{g}SofaGym; computational cost limits interactive use.
\par}
\end{table}

Soft robot simulation spans a wide range of methods (see~\citep{armanini_renda_2023_soft_robot_modeling} for a review), which we organize by their relationship to the requirements in Table~\ref{tab:simulator_comparison}. Approaches based on the finite element method (FEM), such as SOFA~\citep{faure_cotin_2012_sofa} provide high-fidelity deformation and contact modeling~\citep{duriez2013control} but remain too costly for real-time use despite model-order reduction advances~\citep{goury_duriez_2025_towards_realtime}. For slender robots, Cosserat rod models~\citep{rucker_webster_2011_statics_dynamics} reduce the representation from 3D to 1D while remaining geometrically exact, with various strategies achieving real-time free-space dynamics~\citep{till_rucker_2019_realtime_dynamics, renda2014dynamic, renda_seneviratne_2018_discrete_cosserat, renda_boyer_2020_gvs, mathew_renda_2022_sorosim}. However, these are specialized frameworks designed for modeling and analysis: contact on the soft body must be specified as known external loads~\citep{wiese_raatz_2023_describing_contact, xun_kruszewski_2024_cosserat_dynamic}, tendons are approximated as distributed loads rather than discrete forces through tendon paths, and there is no support for complex multi-object scenes or learning pipelines. DisMech~\citep{choi_jawed_2024_dismech} provides physically grounded Kirchhoff rod simulation with robust implicit contact handling and order-of-magnitude speed improvements over Elastica~\citep{gazzola_mahadevan_2018_forward_inverse}, but actuation is limited to natural curvature changes and the framework remains standalone.

Recent work has taken two opposite approaches to bridge this gap. SoftManiSim~\citep{kasaei_mohsen_2024_softmanisim} couples a Cosserat solver with PyBullet, providing physical grounding and learning integration but requiring contact forces on the soft body to be specified as solver inputs rather than discovered by the engine; demonstrated tasks are accordingly limited to trajectory tracking and reaching. SoMo~\citep{graule_wood_2021_somo} instead places soft robots as rigid-link chains directly inside PyBullet, gaining native contact resolution and learning integration, but its parameters are empirically calibrated without continuum-mechanics grounding and its constant-moment actuation assumption breaks down under external loads~\citep{rucker_webster_2011_statics_dynamics, camarillo_salisbury_2008_mechanics_modeling}. As a result, learning demonstrations with these tools remain narrow: free-space reaching for continuum bodies~\citep{kasaei_mohsen_2024_softmanisim} or simple in-hand motions for soft grippers~\citep{graule_wood_2022_somogym}, neither approaching the contact-rich whole-body manipulation we target. Our approach inherits the architectural advantages of the pseudo-rigid-body approach while restoring physical grounding through continuum-informed parameter derivation, enabling sim-to-real learning for contact-rich manipulation.

%===============================================================================
% 3. METHOD
%===============================================================================
\section{Continuum-Informed Rigid-Body Simulation}
\label{sec:method}
We model the TDCR as a Kirchhoff rod (Section~\ref{sec:method_cosserat}), develop a discretization that places it natively inside MuJoCo (Section~\ref{sec:method_discretization}), and outline the sim-to-real pipeline this enables (Section~\ref{sec:method_pipeline}).

\begin{figure}[h]
    \centering
    \vspace*{-2mm}
    \includegraphics[width=13.5cm]{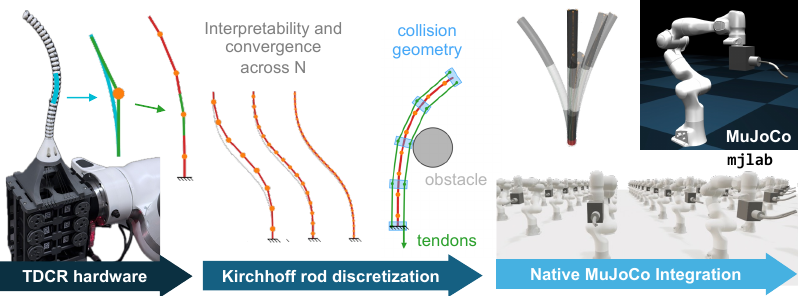}
    \caption{\textbf{Method overview.} Our discretization places a physical TDCR natively inside MuJoCo, with rigid links and elastic joints whose stiffness is derived analytically from material properties. The approximation converges at $O(1/N^2)$. The resulting chain directly leverages MuJoCo's tendon model and contact solver for object interaction. Because the model is encoded in standard MJCF, it is portable across MuJoCo-based engines including parallelized GPU simulation in mjlab; we focus on single-instance MuJoCo simulation in this work.}
\label{fig:method_overview}
\end{figure}

\subsection{Continuum Mechanics Primer}
\label{sec:method_cosserat}

We model a TDCR as a Kirchhoff rod, a special case of the Cosserat rod model~\citep{rucker_webster_2011_statics_dynamics, armanini_renda_2023_soft_robot_modeling} for thin elastic rods. The Cosserat model describes a 3D deformable body as a 1D centerline with a moving cross-section that bends, twists, shears, and stretches along arc length; the Kirchhoff specialization keeps only bending and twisting, which is appropriate for thin TDCRs, whose backbones bend much more easily than they shear or extend (axial and shear stiffnesses $EA, GA \gg$ bending stiffness $EI$). We verify this assumption against a full Cosserat reference in Section~\ref{sec:sim_validation}.

A Kirchhoff rod at each arc-length point $s \in [0, l]$ has three internal degrees of freedom: bending curvatures $(u^x, u^y)$ and torsion $u^z$. These relate to internal elastic moments through the constitutive law
$\bm{\Lambda}_a = \bm{K}_a(\bm{u} - \bm{u}_o)$,
where $\bm{u}_o$ is the rest curvature and $\bm{K}_a = \text{diag}(EI, EI, GJ)$ packages the rod's bending and torsional stiffnesses from elastic moduli ($E$, $G$) and cross-sectional moments ($I$, $J$). The full Cosserat formulation and Kirchhoff specialization are in Appendix~\ref{app:cosserat_full}.

\subsection{Discretization}
\label{sec:method_discretization}

We approximate the Kirchhoff rod as a serial chain of rigid links with elastic joints whose parameters are derived from the rod's material properties. The rod is divided into $N$ segments of length $l_i$, each treated under a piecewise constant strain (PCS) approximation~\citep{renda_seneviratne_2018_discrete_cosserat}: curvatures and torsion are constant within a segment, so under Kirchhoff each segment traces a circular arc. We place a 3-DOF elastic spherical joint at the center of each segment, yielding $N$ joints connected by $N-1$ full-length links plus two half-length links at the base and tip (Figure~\ref{fig:method_overview}). Because each joint sits at its segment's midpoint, the adjacent links are tangent to the circular arc, avoiding the angle compensation and cumulative error propagation seen in other pseudo-rigid-body models~\citep{howell_2001_compliant, santina_rus_2020_model}.
This construction has two key properties (derived in Appendix~\ref{app:derivation}):

\textbf{Physical interpretability:} Applying the constitutive law to each segment gives joint stiffnesses:
\begin{align}
    k_i^x = {EI} / {l_i}, \quad
    k_i^y = {EI} / {l_i}, \quad
    k_i^z = {GJ} / {l_i}.
    \label{eq:stiffness}
\end{align}
These values follow directly from the rod's material and geometric properties. This is in contrast to pseudo-rigid-body models where parameters must be empirically calibrated~\citep{graule_wood_2022_somogym}.

\textbf{Convergence:} The discretization level $N$ is the single free parameter, controlling a direct trade-off between accuracy and computational cost. We fix link lengths to $L_i = l_i/2$ and show in Appendix~\ref{app:convergence} that the static approximation error decreases as $O(1/N^2)$. In practice we use $N = 30$--$50$, where the discretization closely tracks a Cosserat reference at real-time rates (Section~\ref{sec:sim_validation}).

\subsection{Sim-to-Real Pipeline Overview}
\label{sec:method_pipeline}

The discretization above enables us to follow the standard sim-to-real imitation learning workflow for TDCRs: given physical parameters, we generate a MuJoCo model with stiffnesses set by Eq.~\ref{eq:stiffness}; system-identify uncertain parameters against real hardware; collect demonstrations via real-time teleoperation; and train policies for zero-shot deployment. Because the TDCR is a native rigid-body chain, it composes with rigid robots, objects, and environments in a single MuJoCo scene using standard workflows. This is particularly valuable for soft robots: real-time contact simulation supports interactive teleoperation, provides full observability of robot state and contact, and avoids the time and hardware cost of trial-and-error on compliant hardware. We detail each step in Sections~\ref{sec:sim_validation}--\ref{sec:learning}. Because the model is encoded in standard MJCF, it is portable to other MuJoCo-based simulators including mjlab \citep{zakka2026mjlab} for parallelized GPU simulation (Figure~\ref{fig:method_overview}), although we focus on single-instance MuJoCo in this work.

%===============================================================================
% 4. EXPERIMENTS
%===============================================================================
\section{Evaluation}
\label{sec:evaluation}

We evaluate our approach at three levels, finding strong agreement at each: (i) against a Cosserat rod reference, with sub-percent error in statics and dynamics; (ii) against real hardware after system identification, with 4.1\% mean tip tracking error; (iii) and through zero-shot sim-to-real policy deployment, where real-world success rates meet or exceed simulation.

\subsection{Simulation Validation}
\label{sec:sim_validation}

We compare our MuJoCo discretization against SoRoSim~\citep{mathew_renda_2022_sorosim}, a well-established Cosserat rod solver based on the geometric variable strain approach. Since SoRoSim solves the full Cosserat model including shear and extension, this comparison simultaneously validates our discretization and the Kirchhoff assumption for the rod parameters we consider. We test on two materials representing the stiffness range of common TDCRs (Table~\ref{tab:eval_setup}), under both static and dynamic loading, across five discretization levels $N \in \{25, 35, 50, 70, 100\}$.

\begin{wrapfigure}[12]{r}{0.4\linewidth}
  \centering
  \vspace{-90mm}
  \includegraphics[width=\linewidth]{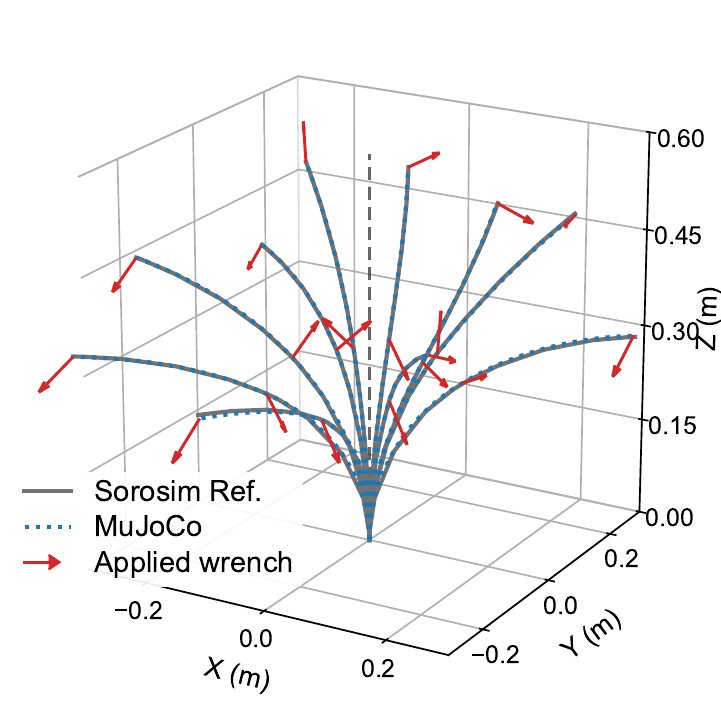}
  \vspace*{-8mm}
  \caption{Representative static test shapes with $N=50$. MuJoCo (dotted) closely matches the
    SoRoSim reference (solid) under randomized gravity
    and applied wrenches (red arrows).}
  \label{fig:statics_shape}
  \vspace{2mm}
  \includegraphics[width=\linewidth]{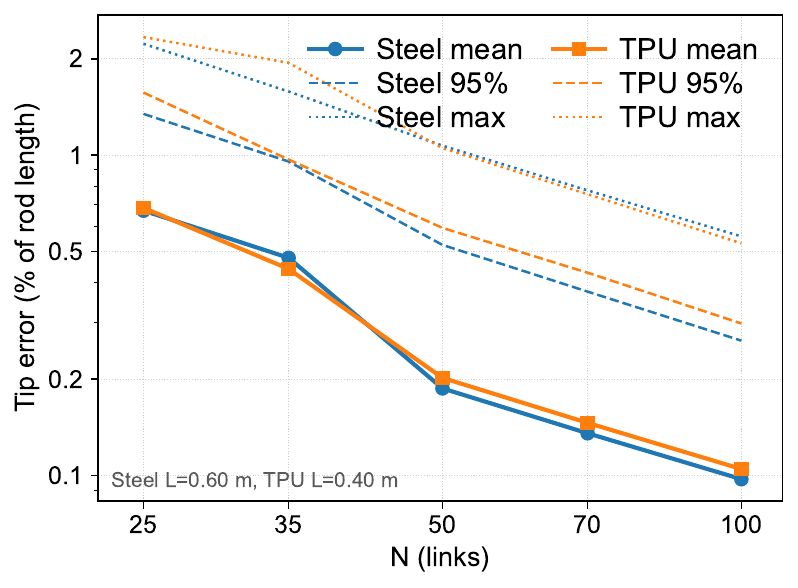}
  \vspace*{-8mm}
  \caption{Static tip error vs.\ discretization level. Mean
  (solid), 95th percentile (dashed), and maximum (dotted) tip error
  as a percentage of rod length, over 500 tests per material.}
  \label{fig:statics_error}
\end{wrapfigure}

\begin{table}[h]
\begin{minipage}{0.5\textwidth}
\small
\caption{Simulation validation setup. Two materials spanning three orders of magnitude in stiffness are tested under randomized loading.}
\label{tab:eval_setup}
\begin{tabular}{@{}lcc@{}}
\toprule
 & Spring Steel & Thermoplastic \\
 & &              Polyurethane (TPU) \\
\midrule
Length $l$              & \SI{0.6}{\meter}  & \SI{0.4}{\meter} \\
Radius $r$             & \SI{0.8}{\milli\meter} & \SI{5}{\milli\meter} \\
Young's mod.\ $E$      & \SI{200}{\giga\pascal} & \SI{70}{\mega\pascal} \\
Poisson's ratio $\nu$  & 0.33 & 0.4 \\
Density $\rho$         & \SI{7870}{\kilo\gram\per\meter\cubed}
                       & \SI{1200}{\kilo\gram\per\meter\cubed} \\
\midrule
\multicolumn{3}{@{}l}{\textit{Static loading (random mid/tip wrench)}} \\
Force (mid/tip)        & \SI{0.20}{\newton} & \SI{0.30}{\newton} \\
Moment (mid/tip)       & \SI{0.08}{\newton\meter} & \SI{0.06}{\newton\meter} \\
\midrule
\multicolumn{3}{@{}l}{\textit{Dynamic (random mid/tip wrench and release at t=0)}} \\
Force (mid/tip)        & \SI{0.40}{\newton} & \SI{0.30}{\newton} \\
Moment (mid/tip)       & 0 & \SI{0.08}{\newton\meter} \\
Damping ratio          & \multicolumn{2}{c}{0.002--0.005} \\
\midrule
Gravity                & \multicolumn{2}{c}{\SI{9.81}{\meter\per\second\squared}, random direction per test} \\
\bottomrule
\end{tabular}
\end{minipage}
\end{table}

\textbf{Static validation.}
For each material, we evaluate on 500 equilibrium shapes under randomized loading: gravity, midpoint wrenches, and tip wrenches are all applied with random 3D directions at fixed magnitudes (Table~\ref{tab:eval_setup}). Figure~\ref{fig:statics_error} confirms that mean tip error decreases consistently with $N$ for both materials, reaching $0.20\%$ of rod length at $N{=}50$ with worst-case errors at $1\%$. The two materials track closely despite spanning three orders of magnitude in mechanical stiffness.

\begin{figure}[h]
\centering
\includegraphics[width=\linewidth]{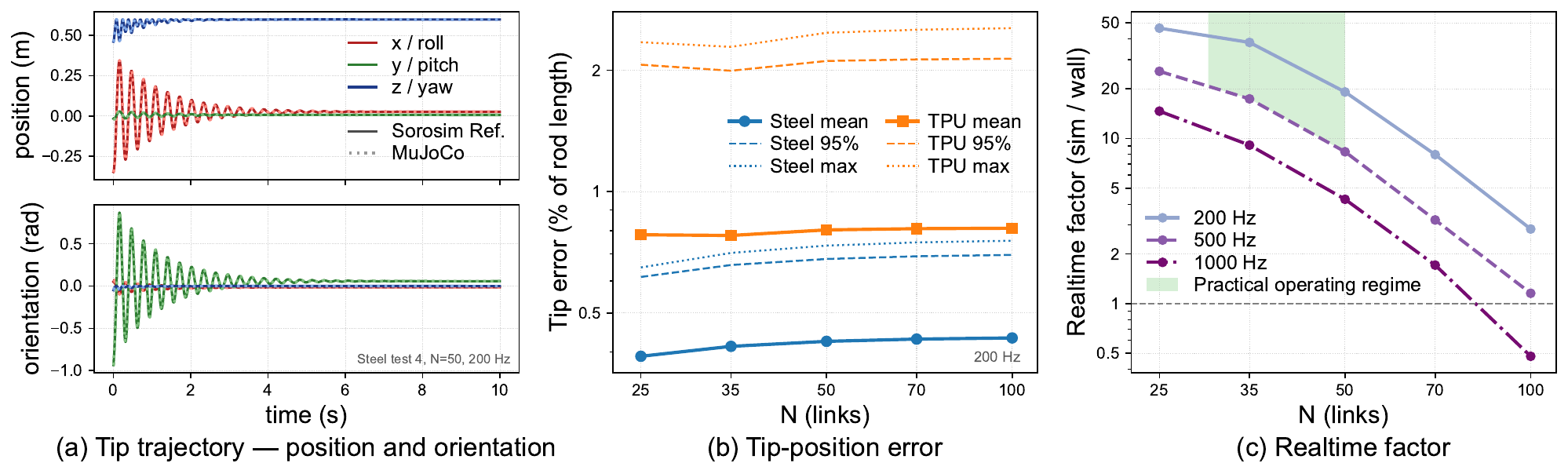}
\caption{\textbf{Dynamic validation against SoRoSim.}
(a)~Tip position and orientation over 10 seconds of free oscillation after wrench release (spring steel, $N{=}50$, 200~Hz).
(b)~Mean tip error vs.\ $N$ for both materials, with 95th percentile and maximum.
(c)~Real-time factor on unoptimized MuJoCo across discretization levels $N$ and simulation rates up to 1000~Hz. The shaded region marks the practical operating regime ($N = 30$--$50$, $\le 500\,\mathrm{Hz}$); dashed line marks real-time (RTF=1).}
\label{fig:sim_dynamic}
\end{figure}

\textbf{Dynamic validation.}
We evaluate dynamic behavior using a tip-release protocol: the rod stabilizes under gravity and applied wrenches, which are then suddenly released at $t=0$, and the rod swings in a damped oscillation until it stabilizes to its rest shape. Across the two materials, we compare 20 such trajectories against SoRoSim. Given the same physical parameters, MuJoCo recovers SoRoSim's oscillation frequency and damping decay exactly (Figure~\ref{fig:sim_dynamic}a); this agreement is consistent across physics rates. Dynamic tip error stays below $0.5\%$ of rod length for steel and $1\%$ for TPU across all $N$ (Figure~\ref{fig:sim_dynamic}b)---lower than in the static case as errors average over the oscillation cycle. TPU's higher residual reflects its more complex loading (force + moment, vs.\ force-only). Figure~\ref{fig:sim_dynamic}c stress-tests computational performance across $N$ up to 100 and simulation rates up to 1000~Hz on unoptimized MuJoCo. Our practical operating regime ($N = 30$--$50$ at $\le 500\,\mathrm{Hz}$, shaded) sits at the accuracy-cost sweet spot: static and dynamic errors are below $1\%$ while real-time is maintained with substantial margin.

\subsection{Hardware Validation}
\label{sec:real_validation}

The simulation validation in Section~\ref{sec:sim_validation} establishes that our discretization accurately approximates Cosserat rod theory for material commonly used in TDCRs. Here, we verify that it can also represent a \emph{physical} TDCR after system identification, bridging the gap from theory to real hardware.

\textbf{System identification.}
We validate against a TDCR with 3 segments actuated by 9 tendons (3 per segment), mounted on a 7-DoF Franka arm. We generate a MuJoCo model with $N{=}30$ segments and initialize joint stiffness from the backbone's nominal material properties (Eq.~\ref{eq:stiffness}). We then collect a calibration dataset tracking tendon configurations and measure the resulting TDCR tip position using an optical tracker (0.2\,mm rated RMSE, Lyra, NDI), followed by Bayesian optimization over stiffness, tendon stiffness, and tendon pre-tension to minimize the discrepancy between simulated and measured tip positions. Because the model starts from physically grounded parameters, the optimization converges from a reasonable initialization and the identified values remain interpretable.

\begin{figure}[h]
  \centering
  \includegraphics[width=13.5cm]{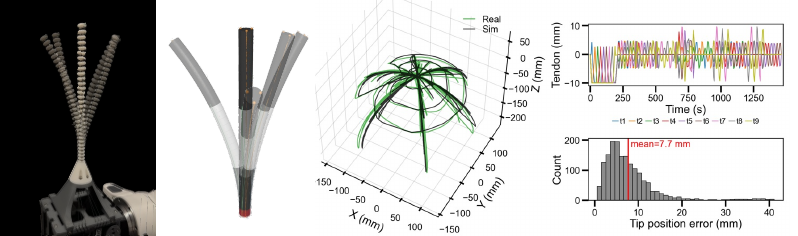}
  \caption{\textbf{Hardware validation.} After system identification,
  the MuJoCo model tracks the physical TDCR with \SI{7.7}{\milli\meter}
  mean tip error ($4.1\%$ of robot length). Left: TDCR prototype and simulation. Centre: Sim-vs-real tip trajectory. Right: Tendon actuation and error distribution across trajectory.}
  \label{fig:real_validation}
\end{figure}

\textbf{Results.}
After system identification, the simulator achieves a mean tip tracking error of ${\sim}$\SI{7.7}{\milli\meter} ($4.1\%$ of the TDCR length of \SI{186.5}{\milli\meter}) across a held-out test set despite various unmodeled effects (e.g., tendon friction, hysteresis, assembly tolerances). Figure~\ref{fig:real_validation} shows sim-vs-real tip trajectory comparisons under varied tendon inputs.

\subsection{Learning Contact-Rich Manipulation}
\label{sec:learning}

\begin{wrapfigure}[8]{r}{0.25\linewidth}
  \centering
  \includegraphics[width=\linewidth]{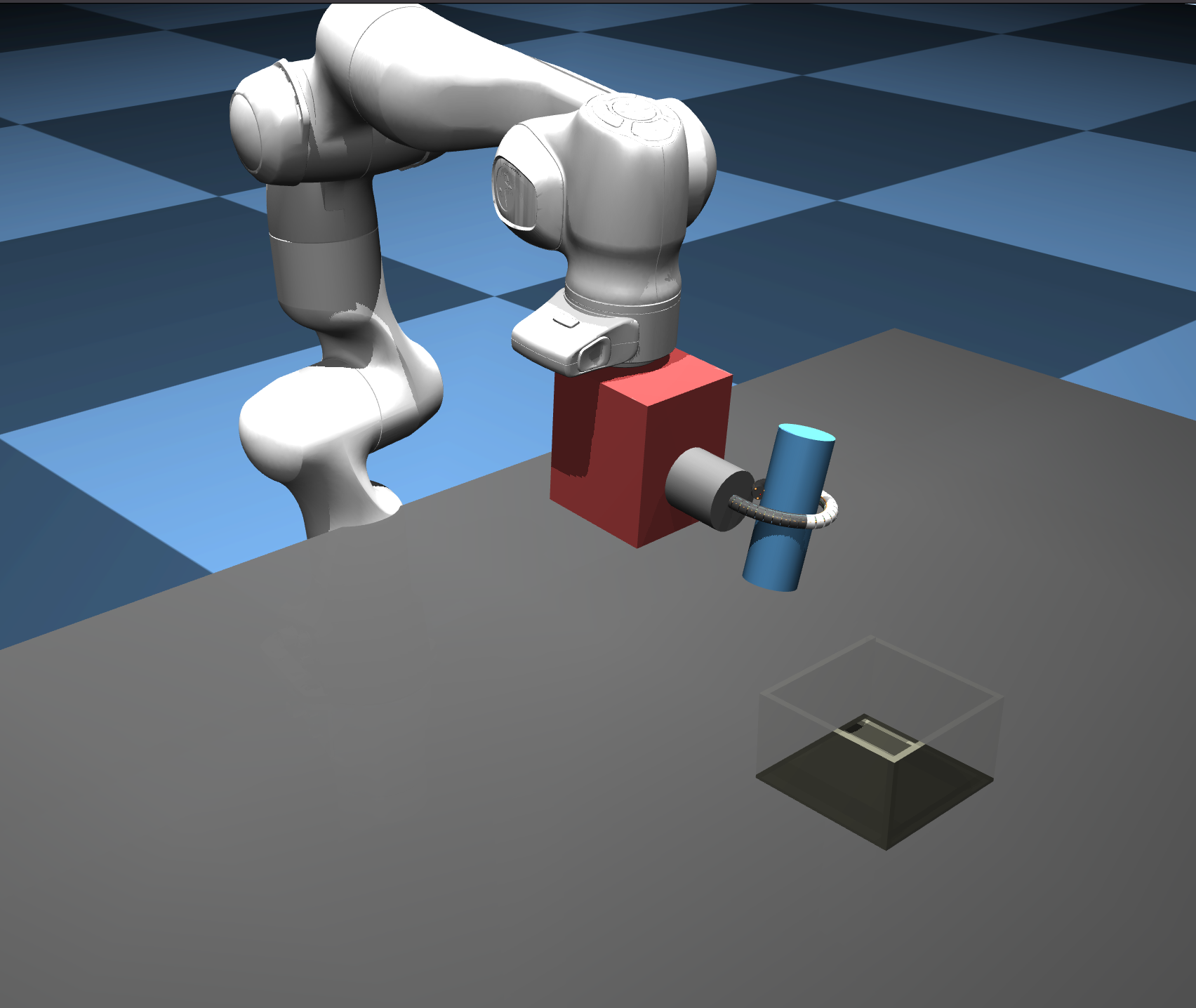}
  \vspace*{-6mm}
  \caption{Grasping task in MuJoCo.}
  \label{fig:cylinder_sim}
\end{wrapfigure}

\begin{wrapfigure}[9]{r}{0.25\linewidth}
  \centering
  \vspace{-2mm}
  \includegraphics[width=\linewidth]{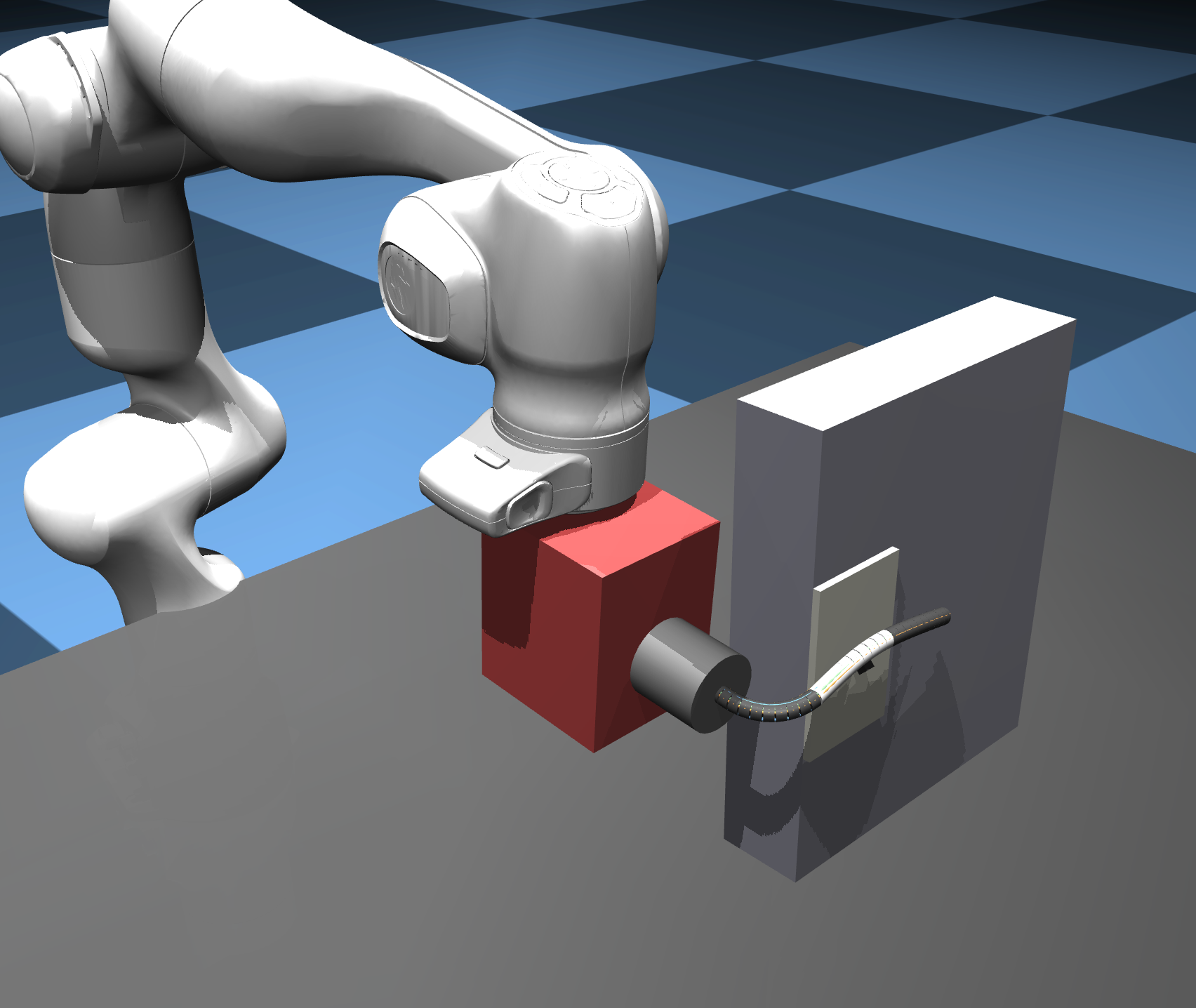}
  \vspace*{-6mm}
  \caption{Switch task in MuJoCo.}
  \label{fig:switch_sim}
\end{wrapfigure}

Our simulation combines real-time performance with both tendon actuation and contact resolution, capabilities not previously available together for soft continuum robots. This allows us to apply the sim-to-real imitation learning workflow (teleoperated data collection, imitation learning, and zero-shot deployment) to a continuum robot, and to test whether the simulator is realistic enough for transfer to succeed. A 9-tendon, 3-segment TDCR is mounted on a 7-DoF Franka arm, with the Franka providing global positioning and the TDCR providing compliant whole-body manipulation. We demonstrate two contact-rich tasks (Figures~\ref{fig:cylinder_sim}, \ref{fig:switch_sim}), both requiring the TDCR to make distributed, unpredictable body contact with the environment to succeed. We use a state-based ACT policy~\citep{zhao_finn_2023_learning} to isolate the contribution of the physics simulation from other factors in sim-to-real transfer (e.g., image-based domain gap).

\textbf{Task 1: Whole-body grasping (Figure~\ref{fig:cylinder_sim}).}
The TDCR must pick up a cylindrical object and drop it into a bin using whole-body contact, wrapping its compliant body around the object rather than grasping with a rigid end-effector.
The policy must coordinate tendon inputs to approach the cylinder from multiple directions in a balanced manner; uneven contact causes the cylinder to tilt or fall before a secure grasp is achieved. Contact is distributed along the body and shifts continuously as the TDCR conforms to the object geometry, an interaction that must be resolved at each timestep and cannot be pre-specified.

\textbf{Task 2: Flip switch from behind (Figure~\ref{fig:switch_sim}).}~%
The TDCR must flip a wall-mounted light switch ($\sim$7 $\times$ 10 mm contact area, $\sim$0.9 N required force) by approaching from behind the wall and pressing downward on the toggle. Soft and compliant robots have traditionally struggled with such precision tasks~\citep{santina_rus_2023_model_based_control} because their distributed flexibility makes it difficult to localize forces accurately on a small target. Compliance itself becomes the solution here: the TDCR braces against the surrounding switch panel to stabilize the body before applying force on the toggle in the correct direction. Failures arise primarily from missing the toggle (misalignment) or applying force at an angle (slipping).

\subsubsection{Data Collection and Training in the Simulator}

\textbf{Observations and actions.}
The policy observes a 19-dimensional state vector: the Franka's 7 joint positions, the TDCR's 9 tendon displacements, and the 3D position of the task target. It outputs a 16-dimensional command of absolute Franka joint and tendon positions. The same I/O is used in simulation and on hardware: at deployment, Franka joint positions are read from FCI, tendon displacements from motor encoders mapped through the drum geometry, and the target position from an NDI Lyra optical tracker with markers attached to the object (Appendix~\ref{app:hardware_sensing}). All observations are normalized before inference.

\textbf{Demonstrations.}
For each task, 50 successful demonstrations are collected via human keyboard teleoperation in MuJoCo. All demonstrations start from the same home configuration (Franka home pose, TDCR with zero tendon displacement; Appendix~\ref{app:training_detail}); only the target object's pose varies between episodes (Table~\ref{tab:CylinderStartingPosition}). After removing consecutive duplicate frames produced by the keyboard interface (44.4\% and 43.3\% of raw frames for the two tasks), demonstrations average ${\sim}9\,\mathrm{s}$ for Task 1 and ${\sim}8\,\mathrm{s}$ for Task 2 (${\sim}500$ frames each at 60\,Hz).

\textbf{Imitation learning.}
We use a state-based variant of ACT~\citep{zhao_finn_2023_learning}, trained for 50{,}000 iterations with the AdamW optimizer (full hyperparameters in Appendix~\ref{app:training_detail}). We do not domain-randomize the simulator's physical parameters: TDCR stiffness, damping, and tendon properties are held fixed at the system-identified values (Section~\ref{sec:real_validation}), and object physics (cylinder mass and friction, switch geometry and force) are manually tuned to real-world values. Randomization is limited to the target object's pose between episodes (Table~\ref{tab:CylinderStartingPosition}) and Gaussian noise added to observations during training to match real-world sensor noise (Table~\ref{tab:ImitationDatasetGaussianNoise}). This minimal randomization lets sim-to-real success be attributed to the simulator's underlying fidelity rather than to randomization compensating for it.

\begin{figure}[t]
  \centering
  \includegraphics[width=\linewidth]{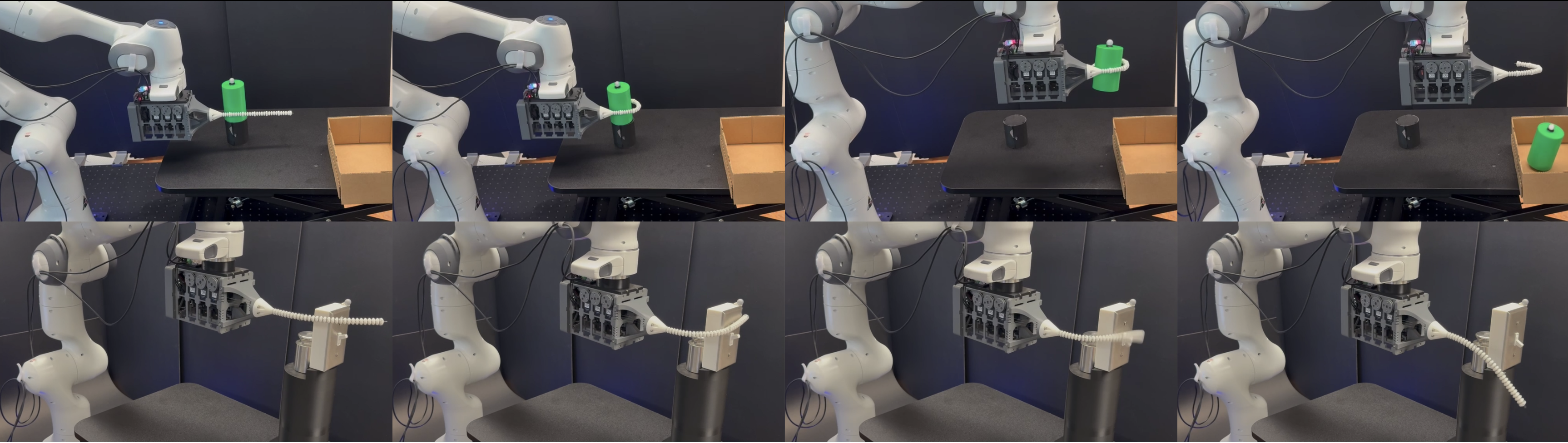}
  \caption{\textbf{Real-world policy rollouts.} All policies trained in simulation and deployed zero-shot. Each task requires emergent whole-body contact: the contact locations and forces arise from the interaction between the policy's actions, the robot's compliance, and the environment geometry.}
  \label{fig:task_rollouts}
\end{figure}

\subsubsection{Sim-to-Real Transfer Results}
\begin{wraptable}{r}{0.4\linewidth}
\vspace{-6mm}
\centering
\caption{Success rates. Policies trained in simulation, deployed
zero-shot on the real system.}
\label{tab:task_results}
\small
\begin{tabular}{@{}lcc@{}}
\toprule
 & Sim & Real \\
\midrule
Grasping & 73\% (73/100) & 76\% (16/21) \\
Switch   & 73\% (73/100) & 78\% (18/23) \\
\bottomrule
\end{tabular}
\vspace{-5mm}
\end{wraptable}
Both tasks were evaluated over 100 randomized episodes in simulation and over 20 trials on real hardware (Table~\ref{tab:task_results}). Real-world success rates slightly exceed simulation rates, a notable outcome for sim-to-real transfer where performance typically degrades from sim to real due to unmodeled physics. We interpret this as evidence that our continuum-informed discretization captures the TDCR's interactions with high enough fidelity. The TDCR's distributed compliance further accommodates small residual errors and environmental variation, an advantage widely noted in the soft robotics literature~\citep{deimel_brock_2016_novel, montero_santina_2024_mastering, chen_2025_survey_adaptability}.

Both tasks achieve near-identical success rates despite stressing different capabilities of the simulator: Task 1 stresses distributed compliant contact while Task 2 stresses precise force application. The consistency across these regimes indicates that performance is bottlenecked by factors common to both (demonstration quality, policy capacity) rather than by deficiencies in the simulator's contact or deformation modeling. Both rollouts qualitatively display the contact-rich manipulation behavior our pipeline is designed to capture (Figure~\ref{fig:task_rollouts}).

%===============================================================================
% 5. CONCLUSION
%===============================================================================
\section{Conclusion}
\label{sec:conclusion}

We show that a continuum-mechanics-informed discretization is sufficient to place soft continuum robots natively inside rigid-body simulators like MuJoCo, enabling a complete sim-to-real pipeline for contact-rich, whole-body manipulation. The discretization is grounded in Kirchhoff rod theory, so we gain contact resolution, tendon routing, and learning integration from existing infrastructure without sacrificing physical interpretability. Building on this, we establish a complete workflow from model generation through teleoperated data collection to zero-shot policy deployment, with real-world success rates that meet simulation across two contact-rich tasks. This brings soft continuum robotics into the simulation-driven learning paradigm that has accelerated rigid-robot manipulation, opening data collection, scaling, and benchmarking workflows that the field has lacked.

\section{Limitations and Future Work}
\label{sec:limitations}

\textit{Scope of validation.}
Our discretization and pipeline are validated on slender, thin-backbone TDCRs where the Kirchhoff assumption holds. Extending beyond Kirchhoff to other soft-robot morphologies, and characterizing MuJoCo's integrator stability and tendon/contact fidelity for soft robotics, are beyond the scope of this work but represent important future directions.

\textit{Closed-loop shape feedback.}
Although the policy closes the loop on tendon positions, the TDCR's full shape (the curvature along its length) is not observed at runtime. Adding such shape feedback would improve robustness under unexpected contact, but real-time shape sensing for TDCRs remains challenging. Vision-based policies that infer shape from observations are a natural next step.

\textit{Learning approach.}
Our work focuses on establishing simulation and system infrastructure rather than advancing the learning algorithm itself. We employ an ACT policy without comparison to alternative architectures or training strategies. Benchmarking against alternative imitation and reinforcement learning methods is an important next step to characterize how learning algorithm choice interacts with simulation fidelity for this robot class.

%===============================================================================
\clearpage
% \acknowledgments{\todo{Funding, collaborators.}}

\bibliography{ref}

%===============================================================================
\clearpage\appendix

%===============================================================================
% APPENDIX A: Full Cosserat Rod Equations
%===============================================================================
\section{Full Cosserat Rod Equations}
\label{app:cosserat_full}

We present the Cosserat rod model in the Lie group formulation following~\citep{renda_seneviratne_2018_discrete_cosserat, boyer2020dynamics}. A Cosserat rod is a one-dimensional continuum whose configuration is described by a field $\bm{g}(s,t) \in SE(3)$ parameterized by arc length $s \in [0, l]$ and time $t$, where $l$ is the rod length in its undeformed state.

\paragraph{Kinematics.}
The spatial and temporal evolution of the configuration are governed by:
\begin{align}
    \bm{g}_s &= \bm{g}\bm{\xi}^\wedge, \label{eq:app_fk} \\
    \bm{g}_t &= \bm{g}\bm{\eta}^\wedge, \label{eq:app_vel}
\end{align}
where $(\cdot)_s$ and $(\cdot)_t$ denote partial derivatives with respect to arc length and time, $(\cdot)^\wedge: \mathbb{R}^6 \to se(3)$ is the hat map from vectors to Lie algebra elements, $\bm{\xi}(s,t) \in \mathbb{R}^6$ is the \emph{strain twist}, and $\bm{\eta}(s,t) \in \mathbb{R}^6$ is the \emph{body velocity twist}. The strain twist decomposes as $\bm{\xi} = (\bm{u}^T, \bm{v}^T)^T$, where $\bm{u} = (u^x, u^y, u^z)$ contains the bending curvatures (about $x$ and $y$) and torsion (about $z$), and $\bm{v} = (v^x, v^y, v^z)$ contains the shear strains (along $x$ and $y$) and axial extension (along $z$). The reference strain for a straight, undeformed rod is $\bm{\xi}_o = (0, 0, 0, 0, 0, 1)^T$.

\paragraph{Compatibility.}
The strain and velocity fields are related by the compatibility equation:
\begin{align}
    \bm{\xi}_t &= \bm{\eta}_s + \ad_{\bm{\xi}}\bm{\eta}, \label{eq:app_compat}
\end{align}
where $\ad$ is the adjoint map of $se(3)$, defined for $\bm{a} = (\bm{a}_1^T, \bm{a}_2^T)^T \in \mathbb{R}^6$ as:
\begin{align}
    \ad_{\bm{a}} = \begin{pmatrix} [\bm{a}_1]_\times & \bm{0} \\ [\bm{a}_2]_\times & [\bm{a}_1]_\times \end{pmatrix},
\end{align}
with $[\cdot]_\times$ the skew-symmetric matrix operator such that $[\bm{a}]_\times \bm{b} = \bm{a} \times \bm{b}$.

\paragraph{Dynamics.}
The balance of momentum along the rod is:
\begin{align}
    \bm{M}\bm{\eta}_t - \ad_{\bm{\eta}}^T\bm{M}\bm{\eta} &= \bm{\Lambda}_s - \ad_{\bm{\xi}}^T\bm{\Lambda} + \bm{F}, \label{eq:app_dynamics}
\end{align}
where $\bm{\Lambda}(s,t) \in \mathbb{R}^6$ is the internal wrench (combining moment and force resultants), $\bm{F}(s,t) \in \mathbb{R}^6$ is the external wrench density (e.g., gravity), and $\bm{M}(s) \in \mathbb{R}^{6 \times 6}$ is the cross-sectional inertia matrix.

\paragraph{Constitutive law.}
A linear constitutive relation relates the internal wrench to strain deviation from the reference configuration:
\begin{align}
    \bm{\Lambda} &= \bm{K}(\bm{\xi} - \bm{\xi}_o), \label{eq:app_constitutive}
\end{align}
where $\bm{K}(s) \in \mathbb{R}^{6 \times 6}$ is the cross-sectional stiffness matrix.

\paragraph{Material matrices for a uniform circular cross-section.}
For a rod of radius $r$, material density $\rho$, Young's modulus $E$, and shear modulus $G$ (with $G = E/[2(1+\nu)]$ for Poisson's ratio $\nu$), the inertia and stiffness matrices decompose as $\bm{M} = \text{diag}(\bm{M}_a, \bm{M}_l)$ and $\bm{K} = \text{diag}(\bm{K}_a, \bm{K}_l)$, with:
\begin{align}
    \bm{M}_a &= \text{diag}\left(\frac{\rho \pi r^4}{4}, \frac{\rho \pi r^4}{4}, \frac{\rho \pi r^4}{2}\right), &
    \bm{M}_l &= \text{diag}\left(\rho \pi r^2, \rho \pi r^2, \rho \pi r^2\right), \label{eq:app_inertia} \\
    \bm{K}_a &= \text{diag}\left(EI, EI, GJ\right), &
    \bm{K}_l &= \text{diag}\left(GA, GA, EA\right), \label{eq:app_stiffness}
\end{align}
where $I = \pi r^4 / 4$ is the second moment of area, $J = \pi r^4 / 2$ is the polar moment, and $A = \pi r^2$ is the cross-sectional area.

\paragraph{Kirchhoff specialization.}
The Kirchhoff rod assumes no shear or axial elongation: $\bm{v} = (0, 0, 1)^T$, i.e., $v^x = v^y = 0$ and $v^z = 1$. This constrains the strain twist to:
\begin{align}
    \bm{\xi} = (u^x, u^y, u^z, 0, 0, 1)^T,
\end{align}
reducing the independent strain components to three: bending curvatures $(u^x, u^y)$ and torsion $u^z$. The arc-length parameter $s$ equals the physical length along the rod (since there is no extension), and each cross-section remains perpendicular to the centerline tangent (since there is no shear).
The internal wrench adopts a decomposition: $\bm{\Lambda} = [\bm{\Lambda}_a, \bm{\Lambda}_l]^T$.
The internal moment $\bm{\Lambda}_a$ satisfies a reduced constitutive law based on Eq.~\eqref{eq:app_constitutive}:
\begin{align}
    \bm{\Lambda}_a &= \bm{K}_a(\bm{u} - \bm{u}_o), \label{eq:ang_constitutive}    
\end{align}
where $\bm{u}=(u^x, u^y, u^z)^T$, and $\bm{u}_o=(0,0,0)^T$ if the reference configuration is straight.
The internal force $\bm{\Lambda}_l$ enforces the Kirchhoff constraints and no longer appears as an independent state.

This assumption is appropriate when $EA \gg EI$ and $GA \gg EI$, which holds for thin rods: $EA/EI = 4/r^2$, exceeding $10^6$ for sub-millimeter backbones typical of TDCRs.

\paragraph{Tendon actuation.}
For tendon-driven continuum robots, tendon forces can be incorporated in two equivalent ways~\citep{rucker_webster_2011_statics_dynamics, boyer2020dynamics,tummers2023cosserat}.
In particular, the effect of multiple tendons can either be modeled as a density of external wrenches $\bm{F}_{\text{tendon}}$ and added to $\bm{F}$~\citep{rucker_webster_2011_statics_dynamics}, or be modeled as a density of internal wrenches $\bm{\Lambda}_{\text{tendon}}$ and added to $\bm{\Lambda}$~\citep{boyer2020dynamics}.

In both formulations, the tendon forces are \emph{distributed} along the rod and depend on the local tendon routing geometry and tendon tension. This is in contrast to the constant-moment assumption used in some pseudo-rigid-body models~\citep{graule_wood_2022_somogym}, where a fixed torque is applied at each joint regardless of configuration.

In our approach, we do not use either continuum tendon formulation. Instead, we leverage MuJoCo's native tendon model, which routes tendons as spatial paths through discrete sites on rigid bodies, with configuration-dependent moment arms and force distribution. This is a natural match for the physical design of TDCRs, where tendons are routed through discrete spacer disks rather than continuously distributed along the backbone.

%===============================================================================
% APPENDIX B: Full Discretization Derivation
%===============================================================================
\section{Full Discretization Derivation}
\label{app:derivation}

We derive the rigid-body approximation of a Kirchhoff rod described in Section~\ref{sec:method_discretization}. The rod is divided into $N$ segments along arc length, each of length $l_i$ with $\sum_{i=1}^N l_i = l$. Within each segment, the strain is assumed constant (PCS approximation).

\subsection{Geometric Construction}

Under the Kirchhoff assumption with constant strain, each segment traces a circular arc in 3D characterized by the segment's bending curvature $\kappa_i = \sqrt{(u_i^x)^2 + (u_i^y)^2}$ and torsion $u_i^z$. We separate the bending and torsion components: the bending defines a planar circular arc, and the torsion defines a rotation about the tangent axis.

For the bending component, consider segment $i$ with constant curvature $\kappa_i$ and arc length $l_i$. The segment subtends an angle $\phi_i = l_i \kappa_i$ at the center of curvature. We place a 3-DOF elastic joint at the \emph{midpoint} of the arc. Two rigid links extend from this joint: one toward the proximal endpoint and one toward the distal endpoint of the segment, each tangent to the arc at the respective endpoint.

By symmetry, each link has equal length $L_i$. From the geometry of a circular arc with radius $R_i = 1/\kappa_i$ and subtended half-angle $\phi_i/2$, the link length is:
\begin{align}
    L_i = R_i \tan\left(\frac{\phi_i}{2}\right) = \frac{\tan(l_i \kappa_i / 2)}{\kappa_i}. \label{eq:app_link_length_tan}
\end{align}
This can equivalently be written using trigonometric identities as:
\begin{align}
    L_i = \frac{\sin(l_i\kappa_i)}{\kappa_i[1 + \cos(l_i\kappa_i)]}. \label{eq:app_link_length}
\end{align}
In the limit $\kappa_i \to 0$ (straight segment), $L_i \to l_i/2$.

The joint angle at the midpoint equals the total bending angle of the segment. Since the two links are tangent to the arc at the endpoints, the angle between them is:
\begin{align}
    \phi_i = l_i \kappa_i.
\end{align}
Decomposing into the two bending directions and torsion:
\begin{align}
    \bm{\theta}_i = l_i \bm{u}_i = (l_i u_i^x,\; l_i u_i^y,\; l_i u_i^z). \label{eq:app_joint_angles}
\end{align}

\subsection{Joint Stiffness Derivation}

The elastic moments at the joint approximate the internal bending and torsional moments of the rod segment. From the constitutive law (Eq.~\ref{eq:app_constitutive}), the bending moment about $x$ in the segment is:
\begin{align}
    \Lambda_{a,i}^x = K_a^x \, u_i^x = EI \, u_i^x.
\end{align}
The corresponding joint torque is $\tau_i^x = k_i^x \, \theta_i^x$, where $k_i^x$ is the joint stiffness and $\theta_i^x = l_i u_i^x$ is the joint angle. Equating:
\begin{align}
    k_i^x \, \theta_i^x = EI \, u_i^x \quad \Longrightarrow \quad k_i^x \, l_i \, u_i^x = EI \, u_i^x \quad \Longrightarrow \quad k_i^x = \frac{EI}{l_i}.
\end{align}
By the same argument for the $y$-bending and torsion:
\begin{align}
    k_i^x = \frac{EI}{l_i}, \quad k_i^y = \frac{EI}{l_i}, \quad k_i^z = \frac{GJ}{l_i}. \label{eq:app_stiffness}
\end{align}
The inverse scaling $k \propto 1/l_i$ ensures that shorter segments (finer discretization) have stiffer individual joints, maintaining the correct total elastic response across the rod. This is a direct consequence of the constitutive law and distinguishes our approach from pseudo-rigid-body models where stiffnesses are empirically chosen.

\subsection{Link Length Mismatch Analysis}

The link length $L_i$ (Eq.~\ref{eq:app_link_length}) differs from $l_i/2$ when the segment has nonzero curvature. We analyze this mismatch.

The derivative of $L_i$ with respect to $\kappa_i$ is:
\begin{align}
    L_i'(\kappa_i) = \frac{l_i \kappa_i - \sin(l_i \kappa_i)}{\kappa_i^2 (1 + \cos(l_i \kappa_i))} \geq 0,
\end{align}
confirming that $L_i$ increases monotonically with curvature: the link must be longer than $l_i/2$ to remain tangent to a curved arc. The mismatch is:
\begin{align}
\label{eq:length mismatch}
    L_i - \frac{l_i}{2} = \frac{\sin(l_i\kappa_i) - l_i\kappa_i\frac{1 + \cos(l_i\kappa_i)}{2}}{\kappa_i(1 + \cos(l_i\kappa_i))},
\end{align}
which converges to 0 as $l_i\to0$ or $N\to\infty$.

There are two approaches to handle this mismatch:
\begin{enumerate}
    \item \textbf{Passive prismatic joint:} Add a prismatic joint to each link whose extension depends on $\kappa_i$ through Eq.~\ref{eq:app_link_length}. This introduces coupling between joint angles and link lengths.
    \item \textbf{Fix $L_i = l_i/2$ and increase $N$:} Accept the mismatch and rely on convergence as $N$ increases. 
\end{enumerate}
We adopt the second approach to maintain compatibility with standard rigid-body simulation tools, which efficiently handle large numbers of fixed-length links. The accumulated mismatch over all links is found to be $O(1/N^2)$, which quickly becomes negligible as $N$ increases (see Section~\ref{app:convergence} for the convergence analysis).

\subsection{Comparison with Prior Discretizations}

\paragraph{Discrete Cosserat approach~\citep{renda_seneviratne_2018_discrete_cosserat}.}
This method also adopts the PCS assumption but constructs the \emph{exact} geometry of each constant-strain segment through spatial integration of the exponential map on $SE(3)$. While this achieves higher geometric accuracy per segment, it requires specialized simulation algorithms (spatial integration at each timestep) and cannot be directly implemented in off-the-shelf rigid-body physics engines. Our approach trades per-segment geometric accuracy for compatibility with general-purpose engines, relying on increased $N$ to recover accuracy.

\paragraph{Augmented rigid-robot model~\citep{santina_rus_2020_model}.}
This method also uses rigid links to approximate continuum robots but is based on the constant-curvature assumption (a special case of PCS without torsion) and enforces only positional matching at segment endpoints. Our PCS-based method additionally captures torsional deformation and matches both position and orientation at segment boundaries through the tangency construction.

%===============================================================================
% APPENDIX C: Convergence Analysis
%===============================================================================
\section{Convergence Analysis}
\label{app:convergence}

We analyze the convergence of our rigid-body discretization to the exact Kirchhoff rod solution as the number of segments $N$ increases. The analysis considers uniform segmentation with $l_i = h = l/N$.

\subsection{Error Sources}

The total approximation error has two sources:

\paragraph{Source 1: Piecewise constant strain (PCS) approximation.}
The continuous strain field $\bm{\xi}(s)$ is approximated by a piecewise constant function $\bm{\xi}_h(s)$ that takes value $\bm{\xi}_i$ on each segment $[s_i,s_{i+1}]$, where $h=s_{i+1}-s_i$. For a smooth strain field $\bm{\xi}\in C^1([0,l])$, this approximation satisfies
\begin{align}
    \|\bm{\xi}-\bm{\xi}_h\|_\infty = O(h),
    \label{eq:app_strain_error}
\end{align}
which is the standard approximation order for piecewise constant interpolation.

The rod configuration $\bm{g}(s)\in SE(3)$ satisfies the kinematic equation
\begin{align}
    \bm{g}_s = \bm{g}\bm{\xi}(s)^\wedge, \qquad \bm{g}(0)=\bm{g}_0.
\end{align}
Over each segment $[s_i,s_{i+1}]$, the exact solution can be written in exponential form using the Magnus expansion:
\begin{align}
    \bm{g}(s_{i+1}) = \bm{g}(s_i)\exp(\Omega_i),
\end{align}
where
\begin{align}
    \Omega_i
    &= \int_{s_i}^{s_{i+1}} \bm{\xi}(s)^\wedge\,ds
    - \frac{1}{2}\int_{s_i}^{s_{i+1}}\int_{s_i}^{\sigma_1}
    \left[\bm{\xi}(\sigma_1)^\wedge,\bm{\xi}(\sigma_2)^\wedge\right]
    \,d\sigma_2\,d\sigma_1 + \cdots .
\end{align}
If $\bm{\xi}$ is sufficiently smooth, then the commutator term is $O(h^3)$, so that
\begin{align}
\label{eq:Omega_i}
    \Omega_i = \int_{s_i}^{s_{i+1}} \bm{\xi}(s)^\wedge\,ds + O(h^3).
\end{align}
In the midpoint PCS approximation, the segment strain is chosen as
\begin{align}
    \bm{\xi}_i := \bm{\xi}\!\left(s_i+\frac{h}{2}\right).
\end{align}
By midpoint quadrature, assuming $\bm{\xi}\in C^2([0,l])$,
\begin{align}
\label{eq:int xi}
    \int_{s_i}^{s_{i+1}} \bm{\xi}(s)^\wedge\,ds
    = h\,\bm{\xi}_i^\wedge + O(h^3).
\end{align}
Combining Eqs.~\eqref{eq:Omega_i} and \eqref{eq:int xi} yields:
\begin{align}
    \Omega_i = h\,\bm{\xi}_i^\wedge + O(h^3).
\end{align}
The PCS integration on segment $i$ is therefore
\begin{align}
    \bm{g}_h(s_{i+1}) = \bm{g}_h(s_i)\exp\!\left(h\,\bm{\xi}_i^\wedge\right),
\end{align}
which has local truncation error $O(h^3)$. Over $N=l/h$ segments, this yields the global tip error
\begin{align}
    d\!\left(\bm{g}(l),\bm{g}_h(l)\right)=O(h^2)=O(1/N^2),
    \label{eq:app_pcs_config_error}
\end{align}
where $d(\cdot,\cdot)$ is a left-invariant metric on $SE(3)$. This is the standard second-order convergence of midpoint Lie group integration~\citep{renda_seneviratne_2018_discrete_cosserat}.

\paragraph{Source 2: Chord-arc (link length) mismatch.}
Our discretization fixes $L_i = l_i/2 = h/2$ rather than using the curvature-dependent expression (Eq.~\ref{eq:app_link_length}). 
Taylor expansion of length mismatch per segment (Eq.~\ref{eq:length mismatch}) gives:
\begin{align}
    L_i - \frac{h}{2} = \frac{\kappa_i^2 h^3}{24} + O(h^5).
\end{align}

Over $N$ segments:
\begin{align}
    \text{Total chord-arc error} = O(N \kappa_i^2 h^3) = O(\kappa_i^2 h^2) = O(1/N^2). \label{eq:app_chord_error}
\end{align}
For bounded curvature ($\kappa_i \leq \kappa_{\max}$), this error is of the same order as the PCS error.

\subsection{Combined Error Bound}

Combining both sources (Eqs.~\ref{eq:app_pcs_config_error} and~\ref{eq:app_chord_error}), the total static approximation error satisfies:
\begin{align}
    d\left(\bm{g}(l), \hat{\bm{g}}_h(l)\right) \leq C \cdot h^2 = \frac{C \cdot l^2}{N^2}, \label{eq:app_total_error}
\end{align}
where $\hat{\bm{g}}_h$ denotes the configuration from our rigid-link discretization (with fixed link lengths) and $C$ is a constant depending on the strain field smoothness and maximum curvature. In terms of the tip position error:
\begin{align}
    \|\bm{p}(l) - \hat{\bm{p}}_h(l)\| = O(1/N^2).
\end{align}

\subsection{Numerical Verification}

We verify the convergence rate empirically using the static validation setup described in Section~\ref{sec:sim_validation}. Figure~\ref{fig:statics_error} plots the mean tip position error against $N$ on a log-log scale. The observed slope is approximately $-2$, confirming the theoretical $O(1/N^2)$ rate. At $N = 70$, the mean tip error is \SI{3}{\milli\meter} ($0.5\%$ of rod length $l = \SI{0.6}{\meter}$), demonstrating that the discretization achieves high accuracy at practical levels.

\subsection{Note on Extension to General Cosserat Rods}

The convergence analysis above relies on two properties specific to Kirchhoff rods:
\begin{enumerate}
    \item The arc-length parameter $s$ equals the physical length along the rod (no extension), so segment boundaries are fixed in the material frame.
    \item Each constant-strain segment traces a circular arc, enabling the tangent link construction.
\end{enumerate}
For general Cosserat rods with shear and extension ($\bm{v} \neq (0,0,1)^T$), neither property holds: the physical length changes with deformation, and constant-strain segments trace more general curves (not circular arcs). Extending the discretization to this case would require: (a) a modified geometric construction that accommodates non-circular segment shapes, and (b) additional translational joints to capture shear and extension, with stiffnesses $k_i^{v_x} = GA/l_i$, $k_i^{v_y} = GA/l_i$, $k_i^{v_z} = EA/l_i$ derived from $\bm{K}_l$. While the stiffness derivation generalizes directly, the geometric construction is a direction for future work.

\section{Hardware System Overview}
\label{app:hardware}

\begin{figure}[h]
    \centering
    \includegraphics[width=13.5cm]{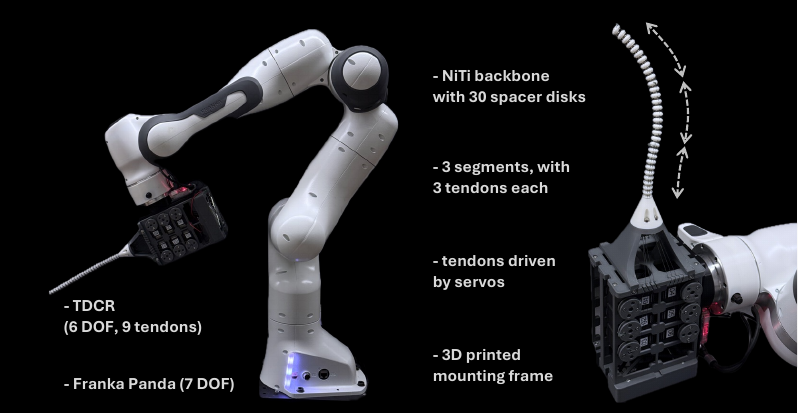}
    \caption{\textbf{Hardware overview.} \textit{(Left)} A 9-tendon, 3-segment TDCR mounted at the end-effector of a 7-DoF Franka Panda arm. The Franka provides global positioning while the TDCR provides distributed compliance for whole-body contact. \textit{(Right)} Close-up of the tendon actuation assembly: nine motors (three per segment) drive tendons routed through spacer disks along the $186.5$\,mm backbone. The actuation housing mounts directly to the Franka flange.}
    \label{fig:hardware}
\end{figure}

The hardware platform combines a Franka Panda 7-DoF arm with a custom 3-segment, 9-tendon tendon-driven continuum robot (TDCR) end-effector (Figure~\ref{fig:hardware}). The Franka provides global positioning while the TDCR provides distributed compliance for whole-body contact.

\subsection{Franka Panda}

We use a Franka Panda arm controlled via a custom Python wrapper built on top of \texttt{libfranka}'s Franka Control Interface (FCI). The wrapper accepts joint position commands at 100\,Hz and linearly interpolates between successive commands to produce the 1\,kHz reference trajectory required by FCI. Safety limits and reachable workspace are configured to match the MuJoCo scene.

\subsection{TDCR Mechanical Design}

\paragraph{Backbone.} The TDCR has a total length of $186.5$\,mm divided into three equal length segments. The backbone is a NiTi rod of diameter 1\,mm. These values serve as the starting point for system identification.

\paragraph{Spacer disks and tendons.} 30 spacer disks of diameter 10\,mm and height 5\,mm are distributed uniformly along the backbone, 10 per segment. Each disk is 3D-printed in PLA with a central hole for the backbone and peripheral holes routing tendons. Tendons are braided Dyneema of diameter 0.5\,mm. Each segment is actuated by three tendons spaced $120^\circ$ around the backbone. Tendons are anchored at the distal disk of their target segment and routed through the spacer disks back to the actuation housing.

\paragraph{Actuation.}
Each of the nine tendons is driven by a Dynamixel XL430-W250-T servo. Tendons are wound onto a drum of diameter 10\,mm fixed to the motor shaft. Motor zero/home positions are calibrated as needed using an optical tracker at the tip and the method described in \citep{lee_burgner_2025_automating}, then commanded with absolute tendon positions relative to home.

\paragraph{Mounting and Interface.}
The actuation housing is rigidly mounted to the Franka flange via a custom 3D-printed adapter plate. The full assembly (housing + TDCR) adds a mass of $\sim$1\,kg, within the Franka's payload limit.

\subsection{Electronics and Control}

The nine motors are daisy-chained over RS485 at 3\,Mbps and interfaced to a host PC. The TDCR control loop runs at 100\,Hz, commanding absolute motor positions. The TDCR and Franka are coordinated at the policy level: at each policy step, the network outputs a 7-DoF Franka joint command and a 9-DoF tendon position command, which are communicated to the respective controllers and tracked independently between policy queries.

\subsection{Sensing}
\label{app:hardware_sensing}
At runtime, the policy observation consists of three components: (i) 7 Franka joint positions from FCI, (ii) 9 motor positions mapped to tendon displacements through the drum geometry, and (iii) the 3D position of the task target, measured by an NDI Lyra optical tracker with markers attached to the manipulated object (cylinder or switch). The TDCR's own shape is not measured at runtime. The same Lyra tracker is also used offline for system identification and validation (Section~\ref{sec:real_validation}), with markers placed at the TDCR tip rather than on the target object.

\section{Imitation Learning Details}
\label{app:training_detail}

The policy used in Section \ref{sec:learning} is a modified ACT architecture ~\citep{zhao_finn_2023_learning}. While the original ACT is designed to accept multiple camera viewpoints, joint positions, and style variables as input, our variant is conditioned solely on joint positions with sensor noise added during training (Table~\ref{tab:ImitationDatasetGaussianNoise}). Hyperparameters are available in Table~\ref{tab:ACTHyperparameters}.

\subsection{Demonstration Success Criteria}
50 successful demonstrations are collected to form the dataset for each task. In all demonstrations the Franka arm begins in the home position specified by Table \ref{tab:FrankaHomeJoints} and with the TDCR in its home position. A successful demonstration is defined differently per task.

The criteria for a successful demonstration of \textbf{Task 1: Whole-body grasping (Figure~\ref{fig:cylinder_sim})} is defined as:
\begin{enumerate}
    \item The cylinder spawns in a random location (Table~\ref{tab:CylinderStartingPosition}) on the table.
    \item The TDCR unit mounted to the Franka end effector is moved near the cylinder on the table.
    \item The TDCR wraps around the cylinder.
    \item The cylinder is lifted off the table by actuating the Franka arm while the TDCR continues to grasp the cylinder.
    \item The cylinder is moved towards the target bin location by actuating the Franka arm while the TDCR continues to grasp the cylinder.
    \item The cylinder is hovering over the target bin and the Franka arm stops actuating.
    \item The TDCR uncurls and releases the cylinder from its grasp.
    \item The cylinder falls into the target bin.
\end{enumerate}

The criteria for a successful demonstration of \textbf{Task 2: Flip switch from behind (Figure~\ref{fig:switch_sim})} is defined as:
\begin{enumerate}
    \item The switch spawns in a random position in the workspace (Table~\ref{tab:CylinderStartingPosition}). The switch is mounted on a plate facing away from the robot. The switch starts in the \textit{on} position pointing upwards. 
    \item The TDCR unit mounted to the Franka end effector approaches the wall from behind.
    \item The TDCR bends towards the switch and makes contact with the plate.
    \item The TDCR bends downwards applying force to the switch lever.
    \item The switch lever is flicked, pointing downwards into the \textit{off} position.
    \item The TDCR and Franka retract from the switch plate.
    \item The demonstration completes when the tip position of the TDCR is behind the wall.
\end{enumerate}

For a demonstration to be considered successful, the operator must perform the above steps in order. For example, if the operator drops the cylinder while transporting it to the target bin, retrieves the fallen cylinder and still drops it into the bin, this demonstration is discarded and not considered successful. If the operator attempts to flick the light switch by applying downward force with the TDCR but misses the switch, the demonstration is discarded and not considered successful. Demonstrations are verified to be successful qualitatively by the operator. 

\subsection{Demonstration Statistics}
After 50 demonstrations are collected for each task, the demonstrations are sanitized to remove consecutive duplicate frames prior to training the ACT policy. Two consecutive frames are considered to be duplicates if robot states are identical within a tolerance of $1 \times 10^{-9}$. Sanitizing demonstrations was empirically found to greatly improve the policy behavior during inference by avoiding freezing behavior mid-task. 

A frame is defined as an observation and action pair. Frames are saved at 60 frames per second while the operator records the demonstration in simulation. In total for \textbf{Task 1}, 49,658 frames were recorded by the operator. 44.4\% of consecutive frames across all 50 observations were duplicates and sanitized, resulting in a final frame count of 27,632. Similarly for \textbf{Task 2}, 44,100 frames were recorded, 43.3\% were found to be consecutive duplicates bringing the final count to 25,003. Most of these duplicates arise from the keyboard teleoperation interface: while the operator pauses between keystrokes (switching between Franka and TDCR or simply to plan the next motion), the full system state holds constant.

Collecting 50 demonstrations took approximately 12-20 minutes wall-clock time for the operator, after some practice. This time includes delays between demonstrations when the simulation resets, completing the task, discarding failed demonstrations and saving data. The 50 demonstrations per task are comparable to those used in the original ACT paper~\citep{zhao_finn_2023_learning} on rigid bimanual tasks, despite our 16-DoF hybrid system and contact-rich task requirements. We interpret this as evidence that high-fidelity simulation reduces the data burden for soft-robot policy learning: the simulator provides clean, fully observable demonstrations without sensor noise or partial observability that physical teleoperation would introduce.

\begin{table}[h]
\centering
\caption{Gaussian noise added to each item in the observation vector prior to training to match hardware noise. Units are radian for Franka joints and meters for other quantities.}
\label{tab:ImitationDatasetGaussianNoise}
\setlength{\tabcolsep}{2.5pt}
\begin{tabular}{@{}lcccccccccc@{}}
\toprule
 Observation & Noise Standard Deviation  \\
\midrule
Franka Arm Joint 1-7
 & 0.01 rad \\
TDCR Tendons 1-9
 &  0.0005 rad \\
 Object position X/Y/Z
 &  0.002 m \\
\bottomrule
\end{tabular}
\end{table}

\begin{table}[h]
\centering
\caption{Randomized target object position during data collection, chosen to span the reachable workspace on the physical test table used at deployment.}
\label{tab:CylinderStartingPosition}
\setlength{\tabcolsep}{2.5pt}
\begin{tabular}{@{}lccc@{}}
\toprule
 & X [m] &  Y [m] &  Z [m] \\
\midrule
Cylinder
 &  $$[0.47, 0.72]$$ &  $$[-0.2, 0.2]$$ &  $$\{0.24\}$$ \\
Switch
 &  $$[0.66, 0.78]$$ &  $$[-0.06, 0.24]$$ &  $$[0.475, 0.55]$$ \\
\bottomrule
\end{tabular}
\end{table}

\begin{table}[h]
\centering
\caption{ACT policy hyperparameters. Where applicable, we note deviations from the original ACT paper ~\citep{zhao_finn_2023_learning}.}
\label{tab:ACTHyperparameters}
\begin{tabular}{lll}
\toprule
\textbf{Parameter} & \textbf{Value} & \textbf{Notes} \\
\midrule
\multicolumn{3}{l}{\textit{Architecture}} \\
Hidden dimension & 512 & Same as paper \\
Attention heads & 8 & Same as paper \\
Feedforward dimension & 2048 & Paper: 3200 \\
Encoder layers & 4 & Same as paper \\
Decoder layers & 1 & Same as paper \\
Latent (VAE) dimension & 32 & \\
Dropout & 0.1 & Same as paper \\
\midrule
\multicolumn{3}{l}{\textit{Training}} \\
Learning rate & $1 \times 10^{-5}$ & Same as paper \\
Weight decay & $1 \times 10^{-4}$ & Same as paper \\
KL weight & 10.0 & Same as paper \\
\midrule
\multicolumn{3}{l}{\textit{Inference}} \\
Chunk size & 75 &  \\
Action steps per query & 1 & Full temporal ensembling \\
Temporal ensemble coeff. & 0.01 & Exponential weighting \\
\bottomrule
\end{tabular}
\end{table}

\begin{table}[h]
\centering
\caption{Absolute joint values of the home position of the Franka Panda arm}
\label{tab:FrankaHomeJoints}
\renewcommand{\arraystretch}{1.4}
\begin{tabular}{@{}cc@{}}
\toprule
\textbf{Joint number} & \textbf{Value (rad)} \\
\midrule
1 & $0$ \\
2 & $-\frac{\pi}{4}$ \\
3 & $0$ \\
4 & $-\frac{3\pi}{4}$ \\
5 & $0$ \\
6 & $\frac{\pi}{2}$ \\
7 & $\frac{\pi}{4}$ \\
\bottomrule
\end{tabular}
\end{table}

\end{document}